\RequirePackage{fix-cm}
\documentclass[preprint,12pt]{elsarticle}
\usepackage{algorithm}% http://ctan.org/pkg/algorithms
\usepackage{algpseudocode}% http://ctan.org/pkg/algorithmicx

\usepackage{amssymb}
\usepackage{float}
\usepackage{hyperref}

\journal{CoRR}

\begin{document}
\begin{frontmatter}



\newtheorem{theorem}{Theorem}

\title{Leapfrogging for parallelism in deep neural networks}


\author{Yatin Saraiya}

\address{847 Moana Court, Palo Alto 94306, CA, USA}
\ead{yatinsaraiya12@gmail.com}

\begin{abstract}
We present a technique, which we term leapfrogging, to parallelize backpropagation
in deep neural networks.  We show that this technique yields a savings of $1 - 1/k$ 
of a dominant term in backpropagation, where $k$ is the number of threads (or gpus).
\end{abstract}


\end{frontmatter}

\section{Introduction}
\label{introduction}
One pass over a neural network consists of 2 phases, forward and backward propagation.
Each phase consists of computations applied at each layer of the neural net, in sequence.  
There are three dominant subcomputations at each level, all matrix computations: of
$z$, $\delta$ and $\nabla w$\footnote{We will use the notation of 
\cite{nielsen} in this paper.}. We present an algorithm, which we call leapfrogging,
to parallelize the computation of $\nabla w$.  The relative speedup in this computation
is $1-1/k$, where $k$ is the number of threads used. 
Our approach seems to be different from existing approaches,
such as pipelining (\cite{pipelining}) and striping (\cite{striping}).  

\section{Computations in one pass}
\label{computations}
We will use the treatment and notation of \cite{nielsen} in this paper.
Consider a neural network with $L$ layers numbered $1,\ldots,L$, in  which each of the hidden layers has
$N$ neurons. The metrics below apply to the hidden layers, although all equations are generally
valid.  

We will use the following notation.  Let $w^l$ be the matrix of weights
at the $l$th layer.  It has dimension $N \times N$.
Let $z^l$ be the vector of weighted inputs to the $l$th layer.  It has dimension $N \times 1$.
let $a^l$ be the vector of activations.  It has size $N \times 1$.
Let $b^l$ be the vector of biases at the $l$th layer.   It has dimension $N \times 1$.
Let C be the cost function for the network. 
Let $\delta^l$ be the vector of
errors at the $l$th layer.  It is of dimension $N \times 1$ for each hidden layer.
Let $\nabla_a^l$ denote the vector of the partial derivatives of the cost with respect to the
activations at the $l$th layer.  Its dimension is $N \times 1$.  
 Let $\sigma$ be the sigmoid function,
and $\sigma^\prime$ be the derivative.  Then  the computation of one pass proceeds as follows,
where $x^1$ is the vector of inputs.
\begin{equation}
\label{x}
a^1 = x^1
\end{equation}
\begin{equation}
\label{z}
z^l = w^{l} \dot a^{l-1}
\end{equation}
\begin{equation}
\label{a}
a^l = \sigma(z^l)
\end{equation}
The above equations define the forward pass.
The following equations apply to backpropagation.
\begin{equation}
\label{bp1}
\delta^L = \nabla_a^L C \odot \sigma^\prime(z^L)
\end{equation}
\begin{equation}
\label{bp2}
\delta^l = ((w^{l+1})^T \delta^{l+1}) \odot \sigma^\prime(z^l)
\end{equation}
\begin{equation}
\label{bp3}
\frac{\partial C}{\partial b_j^l} = \delta_j^l
\end{equation}
We will use $\nabla_b^l$ to denote the vector of $\frac{\partial C}{\partial b_j^l}$.
\begin{equation}
\label{bp4}
\frac{\partial C}{\partial w_{jk}^l} = a_k^{l-1}\delta_j^l
\end{equation}
We will use $\nabla_w^l$ to denote the matrix of $\frac{\partial C}{\partial w_{jk}^l}$.

The dominant computations are Equation~\ref{z}, Equation~\ref{bp2} and
Equation~\ref{bp4}. 

\section{Leapfrogging}
\label{leapfrogging}
The essence of leapfrogging is to create a number of threads, say $k$, so that each thread
computes equations \ref{bp3} and \ref{bp4} at intervals of size $k$ such that the threads are interleaved.
Let the threads be numbered $0, 1, \ldots k-1$, and assume that all quantities have been computed for
the last k layers.  Then for any $j$, the computation by thread numbered $j$ will compute all quantities except for
equations ~\ref{bp3} and \ref{bp4}, and compute these at levels denoted by $mk + j$ for any $m$ such that
$mk+j < L-k-1$.
That is, each thread will compute these equations at only $1/k$th of the layers.  
\begin{figure*}
  \includegraphics[width=0.2\textwidth]{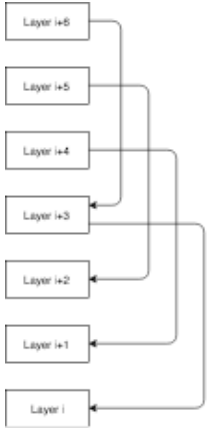}
\caption{Leapfrogging}
\label{bpt}       % Give a unique label
\end{figure*}
Figure~\ref{bpt} shows the picture, where the number of threads $k$ is set to 3.

Algorithm~1 describes the process.
\begin{algorithm}
\label{backpropagation}
\caption{Backpropagation with threads}
\begin{algorithmic}[1]
\Procedure{Backpropagation}{$k$}\Comment{One backward pass. $k$ is the number of threads to use.}
   \State Apply Equation~\ref{bp1} to obtain $\delta^L$
   \For{$i = L-1, L-2, \ldots, L-k+1$} 
	\State Apply equations \ref{bp2}, \ref{bp3} and \ref{bp4} at layer $i$
        \State Save $\nabla_b^i$ and $\nabla_w^i$ to shared memory
   \EndFor
   \State Construct k threads $t_0, t_1, \ldots, t_{k-1}$
   \For{$j= 0, 1, \ldots, k-1$}
       \Call{Thread}{$j$, $k$, $\delta^{L-j}$}
   \EndFor
   \For{$j= 0, 1, \ldots, k-1$}
       \b{join} $t_j$\Comment{Wait for all threads to complete}
   \EndFor
 \EndProcedure
\Procedure{Thread}{$j$, $k$, $\delta$}\Comment{Backward pass with thread $t$ and offset $j$}
   \While{$j < L - k - 1$} 
        \State $i\gets 0$
        \While{$i < k$}  
            \State $l \gets L - 1 - j - i$
	    \State Apply Equation~\ref{bp2} at layer $l$ 
            \If{$i == k - 1$}\label{ifstat}
                \State Apply Equations \ref{bp3} and \ref{bp4} at layer $l$ 
                \State Save $\nabla_b^l$ and $\nabla_w^l$ to shared memory\label{syncstat}
            \EndIf 
            \State $i \gets i + 1$
        \EndWhile
        \State $j\gets j+k$
   \EndWhile
\EndProcedure
\end{algorithmic}
\end{algorithm}
\begin{theorem}[Correctness of Algorithm~1]
\label{correctness}
Algorithm~1 is correct.
\end{theorem}
\paragraph{Proof} 
The parent thread computes Equation~\ref{bp1} at layer L, and 
equations \ref{bp2}, \ref{bp3} and \ref{bp4} for each level $j$
such that $j > L-k$.  
Each child thread computes Equation~\ref{bp2} at every level $j$
such that $j <= L-k$. 
Furthermore, each child thread numbered $j$
computes equations \ref{bp3} and \ref{bp4} at levels $L-1-j-km$, where $m >= 0$, and puts the results in shared memory.
Thus these equations are computed at every layer by some child thread. \qed

\section{Analysis}
\label{analysis}
Our analysis addresses the relative speedup of the entire forward and backward pass.  More precisely,
let $f$ be the total computational cost at any one  layer, and let $f_1$, $f_2$ and $f_3$
be the cost of evaluating equations \ref{z}, \ref{bp2} and \ref{bp4} respectively, sequentially.  Even with the
synchronization cost, it is clear that these three values are dominant, so we write
\begin{equation}
\label{totalsequential}
f = f_1 + f_2 + f_3
\end{equation}
\label{totalthreaded}
Let $f^\prime$ be the cost of Algorithm~\ref{backpropagation}.  It is given by
\begin{equation}
\label{threadsequential}
f^\prime = f_1 + f_2 + f_3/k
\end{equation}
where $k$ is the number of threads.
The relative speedup is then given by
\begin{equation}
\label{speedup}
(f - f^\prime)/f = (1 - 1/k)f_3/f
\end{equation}
Hence the relative speedup for a complete forward pass and backward pass is $1-1/k$ times
the ratio of $f_3$ to $f$, which we will assume is approximately constant for each hidden
layer.  The quantity $1 - 1/k$
rapidly approaches 1 as $k$ increases,  Formally, let $\epsilon$ be the
desired speedup $1-1/k$.  Assume we wish to find the smallest $k$ such that
$1 - 1/k > 1 - \epsilon$.  Then simple algebraic manipulations show that we need to set
the number of threads $k = \lceil 1/\epsilon \rceil$.  However, the absolute speedup depends on the
magnitude of $f_3$.

\section{Compliance}
\paragraph{Conflict of interest}
There are no conflicts of interest on the part of the author.

%\begin{acknowledgements}
%\end{acknowledgements}



\end{document}